\documentclass[11pt]{article}

\usepackage[preprint]{acl}

\usepackage{times}
\usepackage{latexsym}
\usepackage{amsfonts}
\usepackage{amsmath}
\usepackage{subcaption}
\usepackage{graphicx}
\usepackage{pgfplots} 
\usepackage{scalefnt} 

\usepackage{graphicx}
\usepackage{xcolor}
\usepackage{CJKutf8}
\usepackage{tabularx}
\usepackage{booktabs}
\usepackage{makecell}
\usepackage[misc]{ifsym}
\usepackage{amssymb}
\usepackage{pifont}

\usepackage[T1]{fontenc}

\usepackage[utf8]{inputenc}

\usepackage{microtype}

\usepackage{inconsolata}

\usepackage{textcomp}  
\usepackage{scalerel}  
\def\mongochar {\scalerel*{\includegraphics{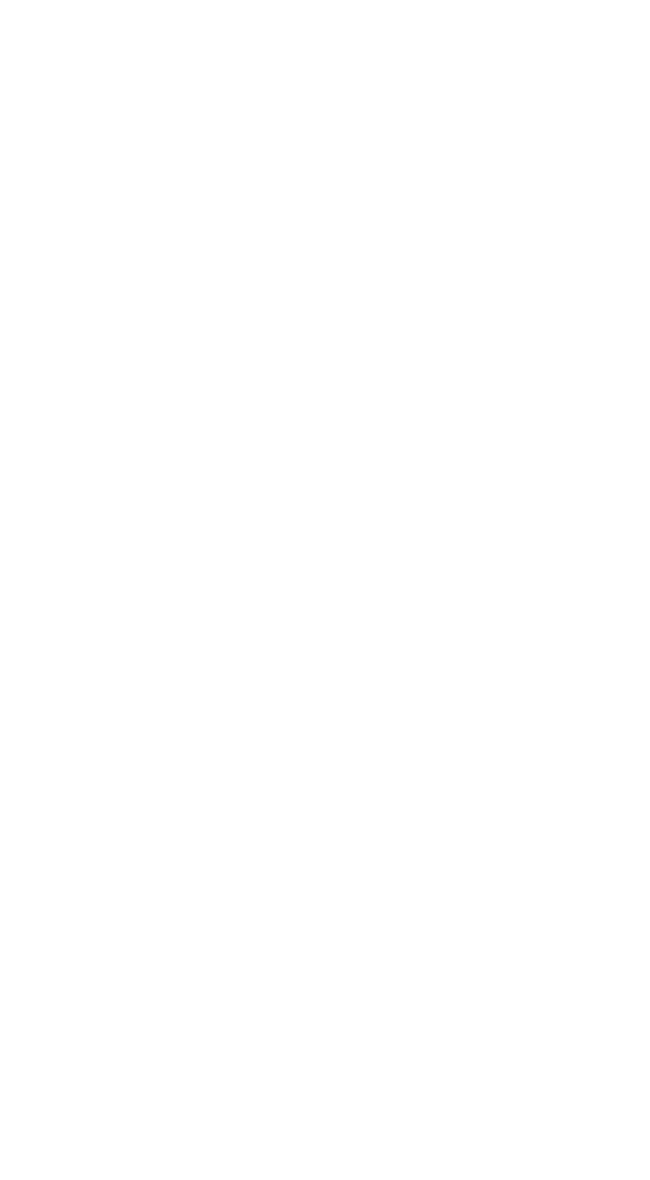}}{\textrm{\textbigcircle}}}
\def\failchar {\scalerel*{\includegraphics{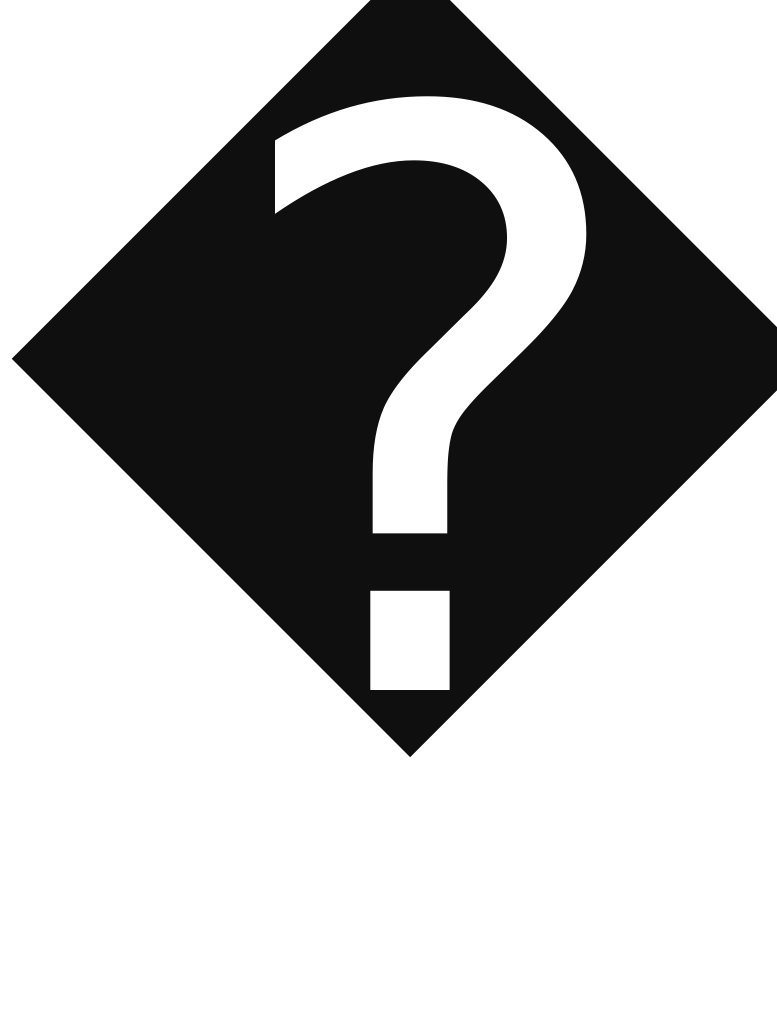}}{\textrm{\textbigcircle}}}
\def\mongoworda {\scalerel*{\includegraphics{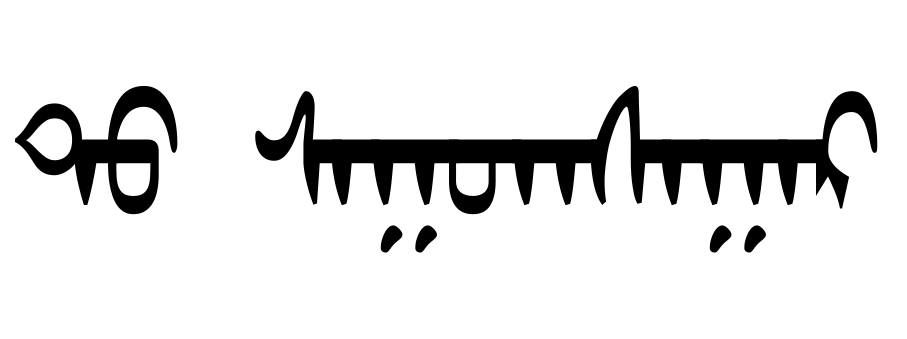}}{\textrm{\textbigcircle}}}
\def\mongowordb {\scalerel*{\includegraphics{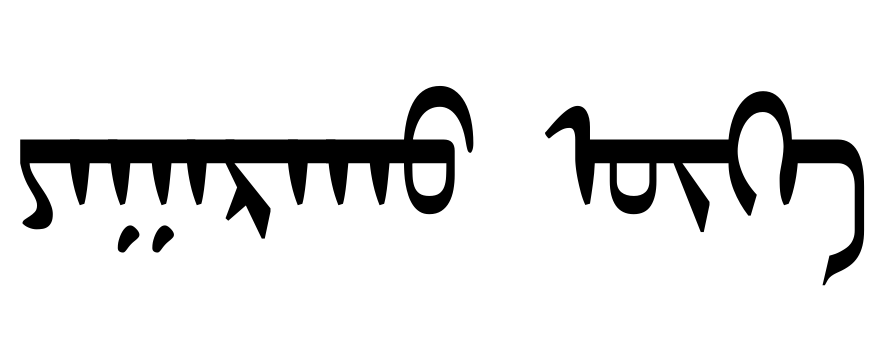}}{\textrm{\textbigcircle}}}

\definecolor{color1}{RGB}{79,79,254}
\definecolor{color2}{RGB}{206,61,50}
\definecolor{color3}{RGB}{175,175,175}
\definecolor{color4}{RGB}{143,80,159}

\newcommand{\red}[1]{\textcolor{red}{#1}}
\newcommand{\blue}[1]{\textcolor{blue}{#1}}
\newcommand{\grey}[1]{\textcolor{color3!70!black}{#1}}

\newcommand{\cyan}[1]{\textcolor{cyan!30!green}{#1}}
\newcommand{\yellow}[1]{\textcolor{yellow!90!purple}{#1}}

\usepackage{multirow}
\usepackage{ulem}

\title{Unlocking the Potential of Model Merging for Low-Resource Languages}

\author{\textbf{Mingxu Tao$^{\text{1,\,3}}$}\thanks{Equal contributions.}, \textbf{Chen Zhang$^{\text{1}}$}\footnotemark[1], \textbf{Quzhe Huang$^{\text{1}}$}\footnotemark[1], \textbf{Tianyao Ma$^{\text{1}}$}\\
\textbf{Songfang Huang$^{\text{4}}$,} \textbf{Dongyan Zhao$^{\text{1,\,2,\,3}}$,} \textbf{Yansong Feng$^{\text{1\,\Letter}}$} \\
$^{\text{1}}$Wangxuan Institute of Computer Technology, Peking University \\
$^{\text{2}}$State Key Laboratory of General Artificial Intelligence, Peking University\\
$^{\text{3}}$Center for Data Science, Peking University \\
$^{\text{4}}$College of Engineering, Peking University \\
{\tt \{thomastao,zhangch,huangquzhe,fengyansong\}@pku.edu.cn} \\
}

\begin{document}
\maketitle
\begin{abstract}

Adapting large language models (LLMs) to new languages typically involves continual pre-training (CT) followed by supervised fine-tuning (SFT). 
However, this CT-then-SFT approach struggles with limited data in the context of low-resource languages, failing to balance language modeling and task-solving capabilities. 
We thus propose a new model merging solution as an alternative for low-resource languages, combining models with distinct capabilities into a single model without additional training. 
We use model merging to develop task-solving LLMs for low-resource languages without SFT data in the target languages. 
Our experiments based on \texttt{Llama-2-7B} demonstrate that model merging effectively endows LLMs for low-resource languages with task-solving abilities, outperforming CT-then-SFT in scenarios with extremely scarce data. 
Observing performance saturation in model merging with increasingly more training tokens, we further analyze the merging process and introduce a slack variable to the model merging algorithm to mitigate the loss of important parameters, thereby enhancing model performance.
We hope that model merging can benefit more human languages suffering from data scarcity with its higher data efficiency.

\end{abstract}

\section{Introduction}

Large language models (LLMs) demonstrate remarkable capabilities across various NLP tasks, owing to the vast amounts of high-quality training data~\cite{touvron2023llama, bai2023qwen}. However, developing models with task-solving abilities for low-resource languages remains challenging due to limited data availability.

A common practice for constructing task-solving LLMs for a low-resource language involves continual pertaining (CT) and supervised fine-tuning (SFT) for the target language~\cite{yong-etal-2023-bloom,nguyen2023seallms}, known as \textbf{CT-then-SFT}.
The scarcity of CT data impedes LLMs' ability to learn effective language modeling for these target languages.
Additionally, it is difficult to acquire sufficient SFT data in low-resource languages to enhance downstream task performance.
To address this issue, previous works attempt to transfer capabilities from high-resource languages to low-resource languages by training on English SFT data~\cite{chirkova2024zero, shaham2024multilingual}.
However, this approach can lead to catastrophic forgetting~\cite{Thrun98, LML18} of language modeling for the target languages~\cite{mehta-2021-et-al-emperical, kotha2024understanding}, resulting in LLMs still failing to solve tasks due to the loss of language abilities.


To better integrate the language modeling and task-solving capabilities, we introduce model merging for low-resource languages, which can combine multiple models with distinct abilities into a single model without additional training.
Previous work~\cite{akiba2024evolutionary} has shown that an LLM for high-resource languages can be merged with task-specific models, such as Japanese language models and math models.
In this work, we explore whether model merging can effectively construct task-solving LLMs for low-resource languages.
Specifically, we investigate the following research questions:
\textbf{RQ1:} What is the viability of constructing task-solving LLMs in low-resource languages through model merging? 
\textbf{RQ2:} Is model merging always a better choice than CT-SFT? 
\textbf{RQ3:} What factors may affect LLMs in obtaining task-solving capabilities through model merging? 

To answer these questions, we study the adaptation of Llama-2-7B~\cite{touvron2023llama}, an English-centric LLM, into seven distinct low-resource languages. We first continually pre-train Llama-2-7B on monolingual texts in each language.
Next, we explore two approaches to inject task-solving capabilities into this continually pre-trained model: (1) training the LLM with English SFT data or the data translated to the target low-resource language; (2) merging the model with an English task-solving LLM.
Experiments show that model merging can effectively equip the CT models with task-solving capabilities. 
Notably, when pre-training corpora in the target language are extremely scarce (<10B tokens), model merging outperforms CT-then-SFT. 
As an LLM is continually pre-trained with more tokens in target languages, the improvements brought by model merging gradually saturate. 
Then, model merging can no longer significantly surpass the SFT method.


To further investigate the factors impeding the continuous improvement of model merging, we conduct a detailed analysis of the process of merging two LLMs. 
We find that when an English SFT model is merged with an LLM continually pre-trained with more tokens in the target language, more parameters from the SFT model are discarded during merging. 
The loss of these parameters may lead to a decline in task-solving capabilities, preventing the merged model from improving performance on downstream tasks.
To mitigate the loss of important parameters from the SFT model, we propose \textit{a novel model merging solution with slack variables}. This strategy allows for more flexible control over the merging process to retain important parameters.


Our contributions are as follows: 
(1) We are the first to introduce model merging to construct task-solving LLMs for low-resource languages;
(2) We reveal that model merging is more effective than SFT in the scenarios of extremely low-resource languages;
(3) Through a quantitative study of the merging process, we explain the performance plateau of model merging with a larger CT corpus and propose a simple yet effective enhancement to popular model merging algorithms.

\section{Related Works}
\paragraph{Model Merging}
Model merging is a promising way to combine the abilities of multiple models. 
Pioneering works explore strategies to find the best weights for averaging~\cite{choshen2022fusing,wortsman2022model,matena2022merging,jin2022dataless}.
Task Arithmetic~\cite{ilharco2022editing} employs task vectors, enabling control 
through arithmetic operations to steer the merged model’s behavior.
TIES~\cite{yadav2023tiesmerging} further addresses the problem of information loss by handling parameter conflict more carefully.
DARE~\cite{yu2023language} zeros out redundant parameters and amplifies the remaining ones.
Evolutionary Model Merge~\cite{akiba2024evolutionary} automatically discovers optimal model combinations through evolutionary algorithms.

There is little discussion of model merging in the context of multilinguality. Instead, previous works attempt to introduce language-specific and task-specific modular adapters~\cite{pfeiffer-etal-2020-mad,parovic-etal-2023-cross,parovic-etal-2024-investigating,zhao2024adamergex}, which require additional training. These works focus on high-resource languages and specific tasks. In contrast, model merging can utilize existing models without additional training, making it a versatile approach for building more general task-solving LLMs. 
Besides, we are the first to study model merging for low-resource languages.













\paragraph{LLMs for Low-Resource Languages}
There is a line of works aiming to adapt LLMs to underrepresented human languages.
A common practice is continually pre-training existing LLMs on the corpus in the target languages~\cite{yong-etal-2023-bloom,nguyen2023seallms,zhang2024mc}. 
To improve the efficiency of training, previous works adopt techniques such as adapters~\cite{pfeiffer-etal-2020-mad}, script conversion~\cite{micallef-etal-2024-cross}, integration of similar languages~\cite{senel-etal-2024-kardes}.

Following pre-training, LLMs typically undergo supervised fine-tuning to acquire task-solving capabilities~\cite{muennighoff-etal-2023-crosslingual,nguyen2023seallms}.
To address the data scarcity in this step, researchers have employed various methods to collect SFT data, including crowd-sourcing~\cite{singh2024aya}, machine translation~\cite{muennighoff-etal-2023-crosslingual,li2023align}, LLM distillation~\cite{li2024x} ,rule-based conversion~\cite{cahyawijaya-etal-2023-instructalign}, et al.
However, these methods are not without limitations, particularly in terms of cost, data quality, and generalizability.
In contrast, the model merging paradigm studied in our work eliminates the need for expensive and potentially error-prone SFT data collection by leveraging pre-trained task-solving models from high-resource languages.





\begin{figure*}[t]
\centering
\includegraphics[width=1\textwidth]{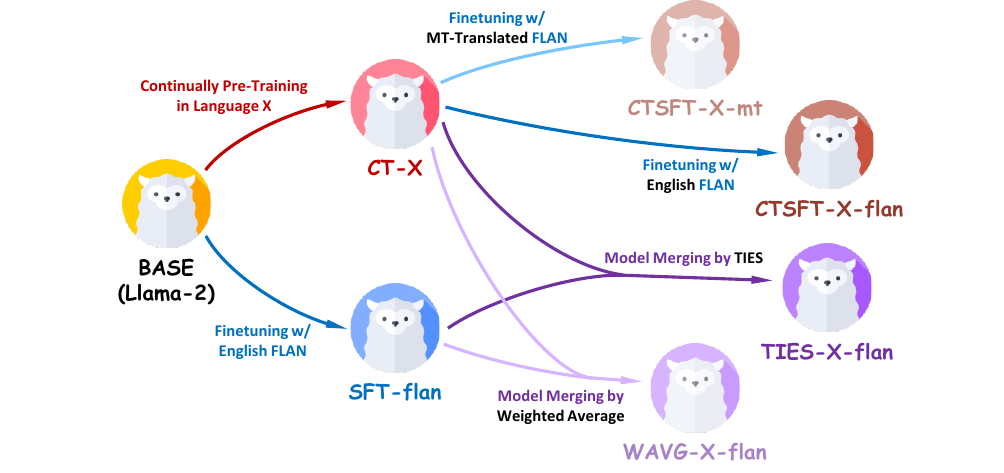}
\caption{Roadmap towards task-solving LLMs for low-resource languages.}
\label{fig:method}
\end{figure*}

\section{Model Merging for Low-Resource Languages} 
The conventional CT-then-SFT paradigm struggles to balance language modeling and task-solving abilities in the context of low-resource languages. 
We propose to adopt model merging as an alternative, which can construct task-solving LLMs for low-resource languages without requiring SFT data in the target languages.

\subsection{Preliminary: Model Merging}
\label{sec:model_merging_intro}

Model merging is a technique for combining multiple models possessing different capabilities into a single versatile model without additional training.
For example, we can merge a model specialized for Japanese and a model specialized for math to obtain a model that excels at solving mathematical problems in Japanese~\cite{akiba2024evolutionary}.
In this work, we investigate two commonly-used methods of model merging: \textbf{weighted averaging}~\cite{choshen2022fusing,wortsman2022model} and \textbf{TIES}~\cite{yadav2023tiesmerging}. 
Here we provide a brief overview of these methods.

\textbf{Weighted averaging} is simply averaging the parameters of two models with a weight tuned on the validation set.

\textbf{TIES} aims to handle the parameter conflicts across multiple models more meticulously. 
Suppose we have two models specialized for distinct tasks, denoted as $\theta_1$ and $\theta_2$, both trained from the same initial model $\theta_\text{init}$.
\textit{Task vectors} for these models are calculated as follows: $\tau_1 = \theta_1 - \theta_\text{init}$ and $\tau_2 = \theta_2 - \theta_\text{init}$. The objective is to merge these 
task vectors and reintegrate them into the initial model. 

The merging process of TIES consists of three steps:
(1) \textbf{Trim}: For $\tau_1$ and $\tau_2$, we trim the redundant parameters by keeping the top-$k_1$\% and  top-$k_2$\% values, respectively, creating $\hat{\tau}_1$ and $\hat{\tau}_2$.  (2) \textbf{Elect Signs}: For each parameter $p$ in $\hat{\tau}_1$ and $\hat{\tau}_2$, we select the sign (+1 or -1) with the higher magnitude, denoted as $\gamma^p = \mathrm{sgn} (\hat{\tau}_1^p +  \hat{\tau}_2^p)$. (3) \textbf{Disjoint Merge}: For each parameter $p$, we only keep the parameter values from $\hat{\tau}_1$ and $\hat{\tau}_2$ whose signs are the same as the aggregated elected sign and calculate their mean.
Specifically, for each parameter $p$, its disjoint mean is calculated as $\tau_{m}^p = \mathrm{avg} (S^p)$, where $S^p = \{ \hat{\tau}_{i}^p | \mathrm{sgn}(\hat{\tau}_{i}^p) = \gamma^p, i = 1, 2\}$.

Given the final merged task vector $\tau_{m}$, we scale it and add it to the initial model $\theta_\text{init}$ to obtain
the merged model $\theta_{m}$ as $\theta_m = \theta_{\text{init}} + \lambda \cdot \tau_m$, where $\lambda$ is a scaling hyperparameter.

For TIES, we tune three hyperparameters in total on the validation set: two sparsity rates $k_1$, $k_2$ and a scaling factor $\lambda$.

\textit{Please refer to the original paper of TIES~\cite{yadav2023tiesmerging} for more details.}

\subsection{Roadmap Towards LLMs for Low-Resource Languages}
\label{sec:roadmap_sft}

Given a base model only pre-trained on an English-centric corpus, e.g., Llama-2-7B~\cite{touvron2023llama} in our study, we want to construct a model capable of solving tasks in a low-resource language. For those target languages, there are usually very limited pre-training texts, ranging from 1B to 20B tokens, and almost no data for supervised finetuning (SFT).
In this scenario, we investigate two representative paradigms of constructing such a model: CT-then-SFT and model merging.
We illustrate the roadmap in Figure~\ref{fig:method}, which demonstrates the relations among the models.

\paragraph{Conventional Practice: CT-then-SFT}
The common practice is (1) first continual pre-training (CT) on the monolingual texts in the target language \texttt{X} to learn the language modeling and (2) then learning task-solving abilities through SFT~\cite{yong-etal-2023-bloom,nguyen2023seallms}. 
This approach is referred to as \textbf{CT-then-SFT}.
Specifically, we consider the following models:

\textbf{BASE}: We employ the original Llama-2-7B without SFT as the base LLM. 

\textbf{CT-\texttt{X}}: We continually pre-train BASE on the corpus in the target language \texttt{X}. Following previous works~\cite{d2019episodic,tao2023can}, we add 1/4 English corpus for memory replay, to avoid catastrophic forgetting English language modeling. 

\textbf{CTSFT-\texttt{X}}: We train CT-\texttt{X} with SFT data to enhance its task solving ability. There are two variants using different SFT data:

(1) \textbf{CTSFT-\texttt{X}-\texttt{flan}}:
We finetune CT-\texttt{X} with English SFT data, which includes the original FLAN datasets and the training set of GSM8K\footnote{
Since the whole instruct-tuning datasets contain over 160K instances, we perform necessary replay with pre-training texts in both English and language \texttt{X}.}.
This approach is based on the assumption that task-solving abilities in English can be transferred to the target language~\cite{chirkova2024zero,shaham2024multilingual}.

(2) \textbf{CTSFT-\texttt{X}-\texttt{mt}}:
We translate FLAN and the training set of GSM8K into the target language \texttt{X} with machine translation (MT) systems\footnote{We use NLLB~\cite{nllb2022} for translation. To enhance the model's ability to follow English prompts, we randomly translate half of the training instances into language \texttt{X}, while the other half of instances remain in English.}, which is a common practice to obtain SFT data for non-English languages~\cite{muennighoff-etal-2023-crosslingual,li2023align,li2023bactrian}. We then finetune CT-\texttt{X} with the obtained SFT data.

\paragraph{New Paradigm: Model Merging} 
By model merging, we can integrate distinct LLMs with various capabilities into one LLM.
To obtain a model capable of solving tasks in the target language \texttt{X}, we can merge the following two models:

\textbf{CT-\texttt{X}}: As discussed in the CT-then-SFT procedure, this model learns a certain amount of language modeling in the language \texttt{X} during CT.
However, its task-solving ability is limited.

\textbf{SFT-\texttt{flan}}: We directly finetune BASE with the SFT data used by CTSFT-\texttt{X}-\texttt{flan}. 
The resulting model has sufficiently learned task solving, but the target language \texttt{X} is still foreign to it.

We merge the two models above to unlock the dual benefits of proficient language modeling and effective task-solving capabilities. 
Specifically, we investigate two methods of model merging: \textbf{weighted averaging} (WAVG, \citealp{choshen2022fusing,wortsman2022model}) and \textbf{TIES}~\cite{yadav2023tiesmerging}. 
We derive two variants of merged models, namely \textbf{WAVG-\texttt{X}-\texttt{flan}} and \textbf{TIES-\texttt{X}-\texttt{flan}}.

\section{Experimental Setup}





\paragraph{Languages}
We use 7 low-resource languages from five distinct language families for experiments: Tamil, Telugu, Odia, Bengali, Tibetan, Uyghur, and Mongolian (in the traditional Mongolian script). 
See their basic information in Table~\ref{tab:language_info}.

We select these languages because they are underrepresented in currently popular LLMs despite their large population (over 475M) worldwide.
As shown in Table~\ref{tab:main_results}, the performance of Llama-2-7B in these languages is close to or even worse than random guessing\footnote{The accuracy of random guessing should be 25\% for Belebele, 14.29\% for SIB-200, and 25\% for the multiple-choice tasks in the MLiC-Eval benchmark.}. 
Notably, the vocabulary of Llama-2 does not even contain tokens for Odia and traditional Mongolian, which indicates that the model has hardly seen these languages during pre-training. 
Moreover, limited resources are available for these languages on the internet.
Among those languages, we can only collect fewer than 1B tokens of monolingual texts for Odia, Tibetan, Uyghur, and traditional Mongolian. 
The problem of data scarcity becomes more severe in terms of high-quality data for supervised fine-tuning.




\paragraph{Pre-training Corpus}
During continual pre-training, we use the largest available corpus for each language from  CulturaX~\cite{nguyen-etal-2024-culturax-cleaned}, IndicCorp-v2~\cite{doddapaneni-etal-2023-towards}, and MC$^2$~\cite{zhang2024mc}.
To maximize language coverage with constraint computational resources, we sample 8B tokens for continual pre-training of Tamil and Telugu, and 16B tokens for Bengali. 
The corpus sizes are shown in Table~\ref{tab:corpus_and_eval}.

Following Llama models~\cite{touvron2023llama}, we employ RedPajama~\cite{together2023redpajama} with the same sampling proportion for memory replay to reduce forgetting of English language modeling. 

\paragraph{SFT Data}
We mainly use FLAN~\cite{longpre2023flan} for SFT, which consists 155K training instances for 1,411 distinct tasks. 
Since there are limited math reasoning tasks in FLAN, we additionally incorporate 7,473 instances from GSM8K~\cite{cobbe2021training} into the supervised training sets.

We translate FLAN into above mentioned languages  using NLLB-200-Distilled-1.3B~\cite{nllb2022}\footnote{NLLB is currently the open-source MT model with the most extensive language support.}. 
Note that this model does not support traditional Mongolian and there are no open-source MT models available for this language currently. We thus adopt a roundabout way: translating the instructions into Cyrillic Mongolian, which NLLB supports, and converting them into traditional Mongolian with an open-source transliteration tool\footnote{\url{https://github.com/tugstugi/mongolian-nlp/tree/master/bichig2cyrillic}}.

\paragraph{Evaluation Tasks}
Regarding the four languages in India (tam, tel, ory, and ben), we use SIB-200~\cite{adelani-etal-2024-sib} for text classification,  Belebele~\cite{bandarkar2023belebele} for machine reading comprehension, and MGSM~\cite{shi2022language} for math reasoning (only available in Telugu and Bengali). 
Regarding the three languages in China (bod, uig, and mvf), we use MLiC-Eval~\cite{mlic_eval}, including the following 4 tasks: text classification (TC), machine reading comprehension (MRC), response selection (RS), and math reasoning.
See statistics in Appendix~\ref{appendix:statistics}.

\paragraph{Implementation Details}
We use Megatron-LM for model training~\cite{shoeybi2019megatron} and Arcee’s MergeKit for model merging~\cite{goddard2024arcee}. 
See more details in Appendix~\ref{appendix:implement}.




\begin{table}[t]
\small
\centering
\setlength\tabcolsep{4pt}
\begin{tabular}{l|ccc}
\toprule
\textbf{Name} & \textbf{Family}  & \textbf{Script} & \textbf{Population} \\
\midrule
Tamil ({tam}) & Dravidian & Tamil & 79M \\
Telugu ({tel}) & Dravidian & Telugu & 96M \\
Odia ({ory}) & Indo-Euro. & Odia & 35M \\
Bengali ({ben}) &  Indo-Euro. & Bengali & 240M \\
Tibetan ({bod}) &  Sino-Tibetan & Tibetan & 7M \\
Uyghur ({uig}) &  Turkic & Arabic & 12M \\
Mongolian ({mvf}) &  Mongolic &  Mongolian & 6M \\
\bottomrule
\end{tabular}
\caption{Languages families, writing systems, and populations of the low-resource languages in our study. }
\label{tab:language_info}
\end{table}

\begin{table}[t]
\small
\centering
\setlength\tabcolsep{3pt}
\begin{tabular}{l|cc|c}
\toprule
\textbf{Lang.} & \textbf{Corpus} & \textbf{Tokens} & \textbf{Tasks} \\
\midrule
tam &  CulturaX & 15.9B & SIB-200, Belebele \\
tel & CulturaX & 12.4B & SIB-200, Belebele, MGSM \\
ory & CulturaX & 765M & SIB-200, Belebele \\
ben & IndicCorp-v2 & 36.4B & SIB-200, Belebele, MGSM \\
bod & MC$^2$ & 1.00B & MLiC-Eval \\
uig & MC$^2$ & 412M &  MLiC-Eval\\
mvf &  MC$^2$ & 904M & MLiC-Eval \\
\bottomrule
\end{tabular}
\caption{The pre-training corpus and evaluation tasks of each language in our study. The number of tokens is obtained with the tokenizer of Llama-2. 
}
\label{tab:corpus_and_eval}
\end{table}

\begin{table*}[t]
\small
\centering
\setlength\tabcolsep{5pt}
\begin{tabular}{l|cc|ccccccc|c}
\toprule
 & \textbf{Task} & \textbf{Lang.} & \textbf{Tamil} & \textbf{Telugu} & \textbf{Odia} & \textbf{Bengali} & \textbf{Tibetan} & \textbf{Uyghur} & \textbf{Mongolian} & \textbf{Average}\\
\midrule
\textbf{BASE} & & & 28.15 & 18.83 & 26.64 & 25.47 & 13.49 & 13.34 & 11.57 & 19.64 \\ 
\textbf{SFT-\texttt{flan}} & \checkmark & & 29.19 & 17.28 & 25.21 & 24.84 & 23.29 & 22.27 & 19.86 & 23.13\\
\textbf{CT-\texttt{X}} &  & \checkmark & 52.18 & 34.67 & 47.93 & 30.77 & 13.52 & 14.80 & 11.09 & 29.28\\ 
\textbf{CTSFT-\texttt{X}-\texttt{mt}} & \checkmark & \checkmark & 50.57 & 32.90 & 30.14 & 38.40 & 33.85 & 24.85 & 19.57 & 32.90 \\
\textbf{CTSFT-\texttt{X}-\texttt{flan}} & \checkmark & \checkmark & 53.95 & \underline{37.96} & 44.56 & \textbf{42.19} & 42.36 & \underline{49.46} & 24.29 & 42.11\\
\textbf{WAVG-\texttt{X}-\texttt{flan}} & \checkmark & \checkmark & \underline{57.56} & 37.58 & \underline{53.59} & 37.19 & \underline{44.30} & 42.64 & \underline{31.09} & \underline{43.42} \\
\textbf{TIES-\texttt{X}-\texttt{flan}} & \checkmark & \checkmark & \textbf{58.46} & \textbf{39.50} & \textbf{56.49} & \underline{40.31} & \textbf{47.86} & \textbf{52.43} & \textbf{32.56} & \textbf{46.80}\\
\bottomrule
\end{tabular}
\caption{Performance of models built through the roadmap. The best results are made \textbf{bold}, with the second \underline{underlined}. The \textbf{Task} and \textbf{Lang.} columns denote whether the model have enhanced task-solving abilities and learned the target languages, respectively. }
\label{tab:main_results}
\end{table*}

\section{Results and Analysis}

\label{sec:main_results}

In this work, we mainly study two categories of common roadmaps, \textit{CT-then-SFT} and \textit{model merging}, to transfer the task-solving ability from a high-resource language to a low-resource one. 
We aim to investigate following research questions.
\textbf{RQ1:} What is the viability of constructing task-solving LLMs in low-resource languages via model merging? 
\textbf{RQ2:} Is model merging always a better choice than CT-then-SFT?  


We take the settings of BASE, SFT-\texttt{flan}, and CT-\texttt{X} as the baselines. 
For \textit{CT-then-SFT}, we investigate two common methods to build task-solving models, CTSFT-\texttt{X}-\texttt{flan} and CTSFT-\texttt{X}-\texttt{mt}. 
For \textit{model merging}, we study two effective algorithms to combine the abilities of language modeling and task solving: weighted averaging and TIES.

Table~\ref{tab:main_results} illustrates the overall performance of different models or setups, i.e., the average scores over all tasks, for each language. 
For the experiments involving continual pre-training, we report the results based on the last checkpoints of CT-\texttt{X}\footnote{Due to the budget of computational resources and available pre-training data, we use 8B tokens for continual pre-training on Tamil and Telugu respectively, and 16B tokens for continual pre-training on Bengali.}. 
See the model performance on individual tasks in Appendix~\ref{appendix:individual_tasks}.

\subsection{Effectiveness of Model Merging}
\label{sec:effectiveness}
For all the studied languages except Bengali, the models constructed by \textit{merging} outperform those built by \textit{CT-then-SFT}\footnote{The exceptional performance in Bengali may be attributed to its larger CT corpus, which we  discuss in Section~\ref{sec:ceiling}.}.
For example, the merged models based on TIES (TIES-\texttt{X}-\texttt{flan}) achieve an average score of 46.80\% across languages, surpassing CT-then-SFT models (CTSFT-\texttt{X}-\texttt{flan}) by +4.69\%.  
Although using a na\"ive merging algorithm, model merging with WAVG (WAVG-\texttt{X}-\texttt{flan}) still achieves better performance than CT-then-SFT (CTSFT-\texttt{X}-\texttt{flan}) in four languages.

In the conventional \textit{CT-then-SFT} approach, LLMs often fail to acquire sufficient comprehension of the target language from a small-size CT corpus, and this ability may be further diminished by supervised fine-tuning. 
In contrast, \textit{model merging} can preserve the language modeling acquired during CT, even with a small-size CT corpus, while incorporating task-solving capabilities by resolving parameter conflicts carefully.

In conclusion, \textbf{model merging can be an effective pathway to obtain task-solving LLMs for low-resource languages}. 

\begin{figure}[t] 
\centering 
\begin{subfigure}{1\columnwidth}
\resizebox{1\columnwidth}{!}{  
    \begin{tikzpicture} 
    \scalefont{0.6} 
    \begin{axis}[
    sharp plot, 
    xmode=normal,
    xlabel=Tokens of CT in Bengali ($\times$1B), 
    ylabel=Average Score (\%), 
    width=8cm, height=3.1cm,  
    xmin=-0.3,xmax=16.5,  
    ymin=28, ymax=44,  
    xtick={0.2,4,8,12,16}, 
    ytick={30,40}, 
    xticklabels={0.2,4,8,12,16}, 
    yticklabels={30,40},
    xlabel near ticks, 
    ylabel near ticks, 
    grid style=dashed, 
    legend style={at={(0.72,0.1)},anchor=south}, 
    legend columns=2, 
    ]

    \addplot+[semithick,mark=*,mark options={scale=0.3}, color=color3] plot coordinates {
        (0.2,31.49) (0.4,31.04) (0.6,30.22) (0.8,30.71) (1.6,30.79) (2.4,30.31) (3.2,29.97) (4,30.85) (4.8,31.25) (5.6,30.92) (6.4,31.41) (7.2,31.21) (8,30.47) (8.8,33.17) (9.6,32.84) (10.4,33.03) (11.2,32.01) (12,33.8) (12.8,31.69) (13.6,31.43) (14.4,32.41) (15.2,31.68) (16,30.77) 
    };
    
    \addplot+[semithick,mark=triangle*,mark options={scale=0.3}, color=color1] plot coordinates { 
        (0.2,36.21) (0.4,37.06) (0.6,37.71) (0.8,38.92) (1.6,38.73) (2.4,39.61) (3.2,40.03) (4,40.46) (4.8,40.22) (5.6,40.76) (6.4,40.18) (7.2,40.3) (8,40.52) (8.8,41.69) (9.6,41.51) (10.4,40.84) (11.2,40.44) (12,40.57) (12.8,40.78) (13.6,40.38) (14.4,40.03) (15.2,40.74) (16,40.31)
    };
    
    \addplot+[semithick,mark=square*,mark options={scale=0.3}, color=color2] plot coordinates {
        (0.2,33.52) (0.4,34.76) (0.6,33.62) (0.8,32.5) (1.6,35.86) (2.4,34.79) (3.2,36.35) (4,36.77) (4.8,37.85) (5.6,36.07) (6.4,36.13) (7.2,37.56) (8,37.85) (8.8,36.97) (9.6,38.56) (10.4,41.32) (11.2,39.78) (12,40.5) (12.8,39.91) (13.6,40.39) (14.4,41.21) (15.2,43.05) (16,42.19)
    };

    \end{axis}
\end{tikzpicture}
}
\end{subfigure}
\vskip 0.1in

\begin{subfigure}{1\columnwidth}
\resizebox{1\columnwidth}{!}{  
    \begin{tikzpicture} 
    \scalefont{0.6} 
    \begin{axis}[
    sharp plot, 
    xmode=normal,
    xlabel=Tokens of CT in Telugu ($\times$1B), 
    ylabel=Average Score (\%), 
    width=8cm, height=3.1cm,  
    xmin=-0.1,xmax=7.5,  
    ymin=31, ymax=41,  
    xtick={0.2,2,4,6}, 
    ytick={32,40}, 
    xticklabels={0.2,2,4,6}, 
    yticklabels={32,40}, 
    xlabel near ticks, 
    ylabel near ticks, 
    grid style=dashed, 
    legend style={at={(0.72,0.1)},anchor=south}, 
    legend columns=2, 
    ]

    \addplot+[semithick,mark=*,mark options={scale=0.3}, color=color3] plot coordinates {
        (0.2,32.56) (0.4,32.66) (0.6,33.39) (0.8,33.53) (1.6,34.29) (2.4,34.82) (3.2,35.08) (4,33.77) (4.8,34.7) (5.6,35.23) (6.4,35.07) (7.2,33.97) 
    };
    
    \addplot+[semithick,mark=triangle*,mark options={scale=0.3}, color=color1] plot coordinates { 
        (0.2,32.8) (0.4,36.35) (0.6,37.02) (0.8,37.54) (1.6,36.49) (2.4,38.29) (3.2,37.54) (4,38.68) (4.8,38.38) (5.6,38.41) (6.4,39.44) (7.2,38.9) 
    };
    
    \addplot+[semithick,mark=square*,mark options={scale=0.3}, color=color2] plot coordinates {
        (0.2,32.54) (0.4,32.89) (0.6,33.88) (0.8,34.3) (1.6,34.17) (2.4,37.12) (3.2,37.02) (4,37.56) (4.8,37.82) (5.6,38.25) (6.4,38.98) (7.2,38.51) 
    };

    \end{axis}
\end{tikzpicture}
}
\end{subfigure}

\caption{Performance of the models based on each checkpoint of CT-\texttt{ben} and CT-\texttt{tel}. The \blue{blue lines} illustrate results of TIES-\texttt{X}-\texttt{flan}, with \red{red lines} for CTSFT-\texttt{X}-\texttt{flan} and \grey{grey lines} for CT-\texttt{X}. } 
\label{fig:performance_detailed_ben_tel}  
\end{figure}
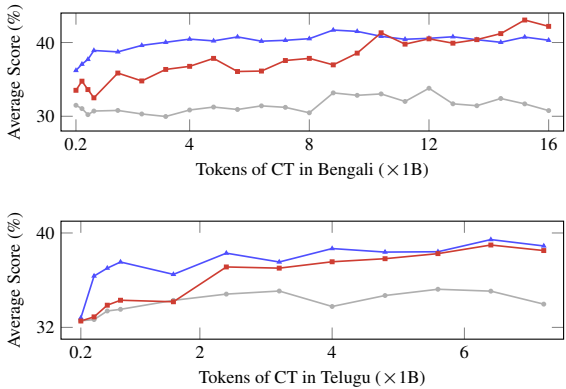

\subsection{Performance Plateau of Model Merging}
\label{sec:ceiling}


Although \textit{model merging} is shown to be more effective than \textit{CT-then-SFT} in almost all languages of our study, this is not the case for Bengali, which has the largest CT corpus in our experiments: CTSFT-\texttt{ben}-\texttt{flan} outperforms TIES-\texttt{ben}-\texttt{flan}.
We guess that the applicability of model merging may be related to the amount of corpus used in CT. 
Thus, we examine the performance changes of the merged models and the CT-then-SFT models under different amounts of CT tokens.

We collect the intermediate checkpoints of CT-\texttt{ben} and CT-\texttt{tel}, which are the two languages with the largest sizes of pre-training corpora. 
Then, we derive CTSFT-\texttt{X}-\texttt{flan} and TIES-\texttt{X}-\texttt{flan} models based on these checkpoints.

Figure~\ref{fig:performance_detailed_ben_tel} illustrates the performance of CT-only (CT-\texttt{X}), CT-then-SFT (CTSFT-\texttt{X}-\texttt{flan}), and model merging (TIES-\texttt{X}-\texttt{flan}) based on every checkpoint. 
For Bengali and Telugu, TIES-\texttt{X}-\texttt{flan} outperforms CT-\texttt{X} in each checkpoint, indicating \texttt{model merging} can robustly enhance LLM's task-solving ability, while \textit{CT-then-SFT} shows greater variability. 
And in both languages, for all checkpoints where the amount of pre-trained tokens is less than 10.4B tokens, 
TIES-\texttt{X}-\texttt{flan} can achieve better results than CTSFT-\texttt{X}-\texttt{flan}. 

\textbf{Model merging can integrate language modeling and task-solving capabilities more effectively than CT-then-SFT, in scenarios with limited language resources~($<$10B tokens in our experiments).}
Four out of seven languages in our study have pre-training corpora with fewer than 10B tokens. This data scarcity is intrinsic for low-resource languages. For instance, 81\% (135 out of 166) of the languages in the multilingual corpus CulturaX~\cite{nguyen-etal-2024-culturax-cleaned} have fewer than 10B tokens. Consequently, our findings on the effectiveness of model merging are broadly applicable and have the potential to benefit a wide range of human languages.


As the amount of CT tokens increases, the CTSFT-\texttt{X}-\texttt{flan} models show more rapid improvement in task-solving capabilities compared to TIE-\texttt{X}-\texttt{flan}. Specifically, when pre-trained with more than 14.4B tokens, CTSFT-\texttt{ben}-\texttt{flan} demonstrates better performance over TIES-\texttt{ben}-\texttt{flan}. 
Similarly, in Telugu, the performance gap between the two models slightly diminishes with additional CT tokens.
However, due to the smaller size of the Telugu corpus compared to Bengali's, we have not observed CTSFT-\texttt{tel}-\texttt{flan} overtaking TIES-\texttt{tel}-\texttt{flan} in our experiments.
We note that \textbf{CT-then-SFT may be a better method to construct task-solving LLMs in languages with sufficient resources, for example, Bengali, Vietnamese, Indonesian, etc.}

\section{Understanding the Dynamics of Model Merging}
\label{sec:dynamics}

As shown in Figure~\ref{fig:performance_detailed_ben_tel}, we find that the performance of TIES-\texttt{X}-\texttt{flan} on downstream tasks may no longer improve as we use more tokens for CT. 
In this section, we want to investigate 
\textbf{RQ3:} What factors may affect LLMs in obtaining task-solving capabilities through model merging?

\begin{figure}[t]
\begin{center}
\centerline{\includegraphics[width=\columnwidth]{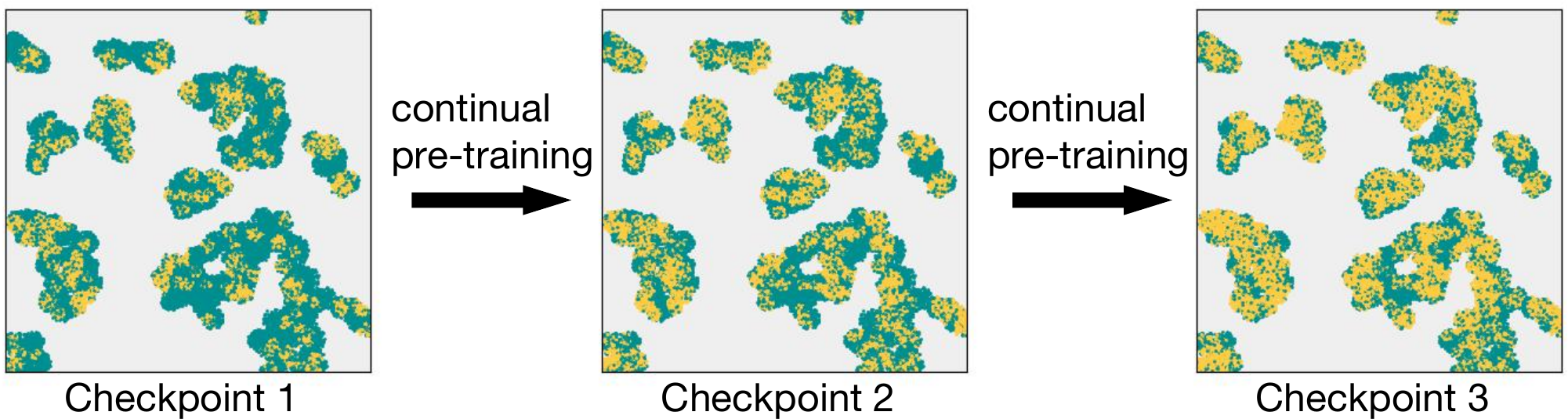}}
\caption{Diagram illustrating the change in the proportion of SFT-\texttt{flan} parameters discarded during sign election as CT progresses. The colored areas represent the parameters with sign conflicts. 
Among them, the \cyan{cyan parts} represent the parameters which elect the signs of SFT-\texttt{flan}, while the \yellow{yellow parts} for parameters selecting the signs of CT-\texttt{ben}.}  
\label{fig:param_disgard} 
\end{center}
\vskip -0.1in
\end{figure}
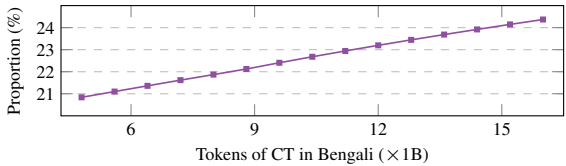
\begin{figure}[t] 
\centering 

\begin{subfigure}{\columnwidth}
\resizebox{1\columnwidth}{!}{  
    \begin{tikzpicture} 
    \scalefont{0.6} 
    \begin{axis}[
    sharp plot, 
    xmode=normal,
    xlabel=Tokens of CT in Bengali ($\times$1B), 
    ylabel=Proportion (\%), 
    width=8cm, height=3cm,  
    xmin=4.3,xmax=16.5,  
    ymin=20, ymax=25,  
    xtick={6,9,12,15}, 
    ytick={21,22,23,24}, 
    xticklabels={6,9,12,15},
    yticklabels={21,22,23,24},
    xlabel near ticks, 
    ylabel near ticks, 
    ymajorgrids=true, 
    grid style=dashed, 
    legend style={at={(0.72,0.1)},anchor=south}, 
    legend columns=2, 
    ]

    
    
    \addplot+[semithick,mark=square*,mark options={scale=0.4}, color=color4] plot coordinates {
        (4.8,20.83904667) (5.6,21.1006621) (6.4,21.3605099) (7.2,21.61810295) (8,21.86930532) (8.8,22.1245605) (9.6,22.40653384) (10.4,22.67888911) (11.2,22.94189952) (12,23.1972387) (12.8,23.44556495) (13.6,23.6858474) (14.4,23.91948392) (15.2,24.14746593) (16,24.36894507)
    };

    \end{axis}

\end{tikzpicture}
}
\end{subfigure}

\caption{
The proportion of parameters discarded in SFT-\texttt{flan} during sign election, relative to the total number of retained parameters of SFT-\texttt{flan} after trimming.} 
\label{fig:magn_avg}  
\end{figure}

\subsection{Quantifying Parameter Conflicts}


Revisiting the mechanism of TIES, we find that during the merging stage, parameters from one model may be discarded due to differences in the parameter signs between the two models. 
As we continually pre-train the LLM with more tokens in language \texttt{X}, the parameters of CT-\texttt{X} changes more significantly.
Since the magnitude of CT-\texttt{X}'s \textit{task vector} becomes larger, more parameters of SFT-\texttt{flan} would be discarded during the process of electing signs.
Figure~\ref{fig:param_disgard} illustrates the changes of discarded parameters in SFT-\texttt{flan}.

The discarding of model parameters usually leads to a decline in the corresponding capabilities. 
Regarding that the CT model's language modeling ability in language~\texttt{X} continuously improves with the increase of pre-training data, we suspect that it is more likely that the SFT model is forced to discard too much information during the merging process, resulting in the merged model's inability to further enhance its task-solving ability in the target language.

To verify this hypothesis, we take Bengali, the language with the largest amount of CT corpus in our study, as an example.
We track the changes in the number of discarded parameters in the SFT model when merging it with CT models trained with different amounts of data. 

As explained in Section~\ref{sec:model_merging_intro}, we first calculate the \textit{task vectors} of SFT-\texttt{flan} and each checkpoint of CT-\texttt{ben}.
In the trimming stage, we find the optimal hyperparameters are $k_{\text{sft}}=0.2$ and $k_{\text{ct}}=1.0$ for most scenarios\footnote{If not being optimal, this set of hyperparameters can also result in comparable performance to the optimal ones.}.
Thus, to provide a fair comparison, we freeze $(k_{\text{sft}},\,k_{\text{ct}})$ as $(0.2,\,1.0)$ for all checkpoints.
Then, we examine the signs and magnitudes of each parameter of the trimmed task vectors of SFT-\texttt{flan} and CT-\texttt{ben}. 
A parameter that has contrary signs in two models can be regarded as a parameter with sign conflict. And if its magnitude in SFT-\texttt{flan} is smaller than that in CT-\texttt{ben}, it will be removed at the stage of sign election.

Figure~\ref{fig:magn_avg} illustrates the proportion of parameters that are discarded from SFT-\texttt{flan} during the merging stage.
We can find as the LLM is pre-trained with more tokens, 
4\% more parameters are removed in trimmed SFT-\texttt{flan}.
We believe that \textbf{when using more tokens for CT, TIES discards a larger proportion of parameters from the SFT model, which may continuously undermine the merged model's task-solving capabilities .}

\subsection{Model Merging with a Slack Variable}


To mitigate the information loss in SFT-\texttt{flan} during the merging stage, we design a new approach,~\textbf{TIES-SV}, enhancing \textbf{TIES} with a \textbf{S}lack \textbf{V}ariable 
to carefully reduce the number of discarded parameters in the model with higher information density, i.e., SFT-\texttt{flan} in this situation.  

According to the process of TIES, for each parameter to be discarded from SFT-\texttt{flan} in the Disjoint Merge step, its magnitude is smaller than the magnitude of its counterpart parameter in CT-\texttt{X}. 
To retain the parameters of SFT-\texttt{flan} while minimizing the information loss of CT-\texttt{X}, we first rank these pairs of parameters between SFT-\texttt{flan} and CT-\texttt{X} according to their differences in the magnitude.
Next, we select a subset of parameters with the smallest magnitude differences to reserve. 

We evaluate the effectiveness of TIES-SV on the model merging process of SFT-\texttt{flan} and the last checkpoint of CT-\texttt{ben}. 
Based on the results in Figure~\ref{fig:magn_avg}, we reserve 4\%  parameters in SFT-\texttt{flan} that were to be discarded in the original algorithm of TIES.
Table~\ref{tab:new_ties} shows the results of vanilla TIES and our TIES-SV on three Bengali tasks. 
In all three tasks, our TIES-SV outperforms vanilla TIES by 0.52\% on average.



\begin{table}[t]
\small
\centering
\setlength\tabcolsep{4pt}
\begin{tabular}{l|ccc|c}
\toprule
 & \textbf{MGSM} & \textbf{SIB-200} & \textbf{Belebele} & \textbf{Average}\\
\midrule
TIES &  7.50 & 79.41 & 34.02 & 40.31 \\
TIES-SV & \textbf{8.00} & \textbf{79.90} & \textbf{34.58} & \textbf{40.83} \\
\bottomrule
\end{tabular}
\caption{Results of TIES and TIES-SV on Bengali tasks.}
\label{tab:new_ties}
\end{table}

It is surprising that this simple and intuitive strategy can bring such an improvement for the merged model, suggesting that different models' parameters may have different importance during the process of model merging. 
\textbf{When parameter conflicts occur, we cannot simply rely on the magnitude of the vectors to decide which models' parameters should be reserve. Instead, one should use prior information obtained through pilot studies or other means to reserve the parameters of the more important model.}
We hope our TIES-SV can shed light on the study of new model-merging algorithms in the future.



\section{Discussions}

In this section, we further explore two important questions related to model merging.
First, we investigate the potential of merging more than one low-resource language into a task-solving model, which could offer a new avenue for constructing multilingual models.
Second, we examine why commonly-used machine-translated SFT data often fails in the context of constructing task-solving models in low-resource languages. 
This failure underscores the advantage of model merging, as it does not require SFT data in the target language.

\subsection{Can We Merge Multiple Languages?}

Previous work~\cite{akiba2024evolutionary} shows that model merging algorithms can be used to combine multiple LLMs with different capabilities, which may enhance the model to solve complex problems.
We wonder whether multiple LLMs adapted to different low-resource languages can be merged with the same SFT model, to construct a task-solving LLM supporting these languages simultaneously.

As a pilot study, we attempt to merge Mongolian and Uyghur LLMs with the English task-solving model SFT-\texttt{flan}. 
Due to the limited budget for computational resources, we cannot conduct a grid search across the hyperparameter space of the three models. 
Therefore, we employ the optimal hyperparameters derived from merging each of the two CT models with SFT-\texttt{flan}.

\begin{table}[t]
\small
\centering
\begin{tabular}{lccc}
\toprule
\textbf{Model} & \textbf{Ability}& \textbf{mvf} & \textbf{uig}\\
\midrule
SFT-\texttt{flan} & \textit{A} & 19.86 & 22.27 \\
CT-\texttt{mvf} & \textit{B} & 11.09 & 13.58\\
CT-\texttt{uig} &\textit{C} & 11.65 & 14.80 \\
\midrule
TIES-\texttt{mvf}-\texttt{flan} & \textit{A+B}& \textbf{32.56} & 27.24 \\
TIES-\texttt{uig}-\texttt{flan} & \textit{A+C}& 24.45 & \textbf{52.43} \\
\midrule
TIES-\texttt{mvf\&uig}-\texttt{flan} & \textit{A+B+C} & \underline{30.61} & \underline{52.40} \\
\bottomrule
\end{tabular}
\caption{Results of merging multiple languages into a task-solving model. The columns \textbf{mvf} and \textbf{uig} refer to the average performance across the Mongolian and Uyghur tasks, respectively.  The best results are made \textbf{bold}, with the second \underline{underlined}. }
\label{tab:model_merge_tri_llm}
\end{table}

Table~\ref{tab:model_merge_tri_llm} illustrates the average scores of tasks in the two languages. 
The merged model serving two low-resource languages (TIES-\texttt{mvf\&uig}-\texttt{flan}) performs comparably to the two single-language merged models (TIES-\texttt{mvf}-\texttt{flan} and TIES-\texttt{uig}-\texttt{flan}) on tasks in the respective languages.

This indicates that \textbf{model merging has great potential for constructing multilingual task-solving LLMs}. We hope this approach can assist multilingual speakers, particularly those using underrepresented languages, by combining multiple existing LLMs in distinct languages without the need for expensive pre-training.



\subsection{Why do Machine Translated Data Fail?}
Collecting synthetic SFT data through MT is an intuitive method for constructing task-solving LLMs in non-English languages~\cite{muennighoff-etal-2023-crosslingual,li2023align}.
However, the experimental results in Table~\ref{tab:main_results} show that MT-translated SFT data may not work when it comes to low-resource languages. 
The models trained with MT-translated data (CTSFT-\texttt{X}-\texttt{mt}) exhibit inferior performance across all languages compared to those trained on English SFT data (CTSFT-\texttt{X}-\texttt{flan}), with a gap of -3.38\%$\sim$-24.61\%.

The decline in model performance can be attributed to the low-quality MT results.
Current open-source MT systems have limited abilities for low-resource languages.
For example, we ask Uyghur and Tibetan native speakers to evaluate sampled translation results by NLLB. They find that there are often irrelevant contents and undesired code-switching in the translation results, as shown in Appendix~\ref{appendix:case_study}.
This kind of noise not only hinders the model's ability to learn task-solving strategies but also interferes with language modeling, which itself is inadequately learned.

In contrast, model merging eliminates the need to collect SFT data in the target languages.
This approach also eliminates the reliance on unreliable machine translation systems.

\section{Case Study}

We further investigate how model merging can improve the model performance on the tasks in low-resource languages. 
Here, we focus on two types of errors in low-resource language models: \textbf{unparsable outputs during generation}, and \textbf{misunderstanding of character overlap} in MRC. 
In this section, we take traditional Mongolian as an example, which the BASE model has hardly seen during pre-training.

\subsection{Unparsable Output}

Note that the tokenizer of Llama-2 does not contain any tokens in the traditional Mongolian script. 
Therefore, the Mongolian texts should be encoded as byte tokens.
For example, the Mongolian character "\mongochar" is tokenized as \texttt{[0xe1, 0xa0, 0xa4]}.
During the inference process, the model generates byte by byte, which can be converted into Mongolian characters by UTF-8.

We find that in the Math task of MLiC-Eval, the models sometimes output incorrect byte combinations. 
These bytes cannot be converted into meaningful UTF-8 characters (displayed as \failchar\failchar). Furthermore, after such errors, the models continue to produce meaningless bytes and fail to provide the final answers.

We find CTSFT-\texttt{mvf}-\texttt{flan} produces meaningless bytes in 74 out of 200 instances, whereas TIES-\texttt{mvf}-\texttt{flan} outputs unparsable bytes in only 42 instances. 
Compared to the CT-then-SFT models, \textbf{the merged models have better abilities to generate texts in low-resource languages.}

\subsection{Misunderstanding of Character Overlap}

We also present a case study of the machine reading comprehension task. 
Here is the English translation of an instance in the MRC task of MLiC-Eval:

\begin{small}
\textbf{Context:}

\ \ Man: Is the train station far from here?
  
\ \ Woman: It only take 20 minutes by taxi.

\textbf{Question:} What does the woman want to express?

\textbf{Choices:} A. Don’t worry. B. It is not too far.
\end{small}

In this instance, the model needs to infer that a 20-minute taxi ride is not too far. 
TIES-\texttt{mvf}-\texttt{flan} can correctly answer this question. 
However, CTSFT-\texttt{mvf}-\texttt{flan} chooses the Choice A. 
This may be due to the character overlap between the phrases "taking a taxi (\mongoworda)" and "don't worry (\mongowordb)" in traditional Mongolian.

The CT-then-SFT model demonstrates worse reasoning capabilities and tends to predict based on character overlap. We guess that fine-tuning on English datasets might primarily obtain the capabilities to handle English tasks but fail to transfer them to the low-resource languages.

\section{Conclusion}
In this paper, we investigate the potential of using model merging to construct task-solving LLMs for low-resource languages. Our findings demonstrate that model merging outperforms the conventional CT-then-SFT paradigm, achieving higher data efficiency. We further analyze the mechanism behind the performance saturation of model merging with an increased number of CT tokens, which inspires a simple yet effective improvement to the model merging algorithm. We hope that model merging can reduce the costs associated with data collection and model training, benefiting a greater number of languages suffering from data scarcity.

\section*{Limitations}
\paragraph{Studied Languages}
Due to the high computational cost of continually pre-training LLMs, we cannot cover a wider range of low-resource languages, only focusing on seven underrepresented languages in India and China. 
However, we make efforts to improve the diversity of selected languages by including different language families and writing systems.

But we have to admit that, for these languages, we can obtain enough language resources to continually per-train LLMs with 7B parameters. 
We still cannot know whether CT-then-SFT and model merging can be effective to build task-solving models in the extremely low-resource languages.
We believe it is important to explore the optimal method to build task-solving LLMs in a wider range of languages.

\paragraph{Evaluation Tasks}
Most of our evaluated tasks focus on natural language understanding, with less emphasis on natural language generation. This limitation arises from the insufficient CT corpus available for the studied low-resource languages, which is insufficient for the models to learn to perform complex generation in the target language.

\paragraph{Model Merging Algorithms} We primarily discuss one popular model merging approach, TIES~\cite{yadav2024ties}, in this work. 
TIES is one of the most effective and efficient methods of model merging yet, which also yields optimal performance in previous work~\cite{akiba2024evolutionary} and does not need to access the training data during merging.
We hope our insights can inspire more studies on other model merging methods in the context of building task-solving LLMs in low-resource languages.

\section*{Acknowledgements}
This work is supported in part by NSFC (62161160339) and Beijing Science and Technology Program (Z231100007423011).
We thank the anonymous reviewers for their helpful discussions and suggestions. For any correspondence, please contact Yansong Feng.

\bibliography{anthology,custom}

\clearpage

\appendix
\section{Data Statistics}
\label{appendix:statistics}
In Table~\ref{appendix:statistics}, we report the statistics of the evaluation datasets used in our study.
We use the development set to tune the hyperparameters in the model merging algorithms and test the models on the test set.

We follow the license for the data used in our work.
Our use of existing datasets is consistent with their intended use.

\section{Implementation Details}
\label{appendix:implement}

Since CulturaX and MC$^{\text{2}}$ are both cleaned and deduplicated corpus, we do not additionally preprocess these data.
In this work, we employ Megatron-LM~\cite{shoeybi2019megatron} for continual pre-training and supervised fine-tuning. 
To obtain the LLMs adapted to each target language, i.e., the CT-\texttt{X} model mentioned in Section~\ref{sec:roadmap_sft}, we continually pre-train Llama-2-7B with the texts in corresponding language. 
Here we use AdamW~\cite{loshchilov2017decoupled} as the optimizer, with $\beta_{1}$ and $\beta_{2}$ are set to $0.9$ and $0.95$ respectively.
The maximum learning rate is set to 2e-5, and the batch size is set to 1M tokens.
We also use bfloat16 to train our models.

For model merging, we employ Arcee's MergeKit~\cite{goddard2024arcee} to merge the CT model in target language and the English SFT model. Following previous works~\cite{yadav2023tiesmerging}, we use grid search~\cite{liashchynskyi2019grid} to select the optimal hyperparameters. 
For the density of each LLM, i.e., $k_1$ and $k_2$ mentioned in Section~\ref{sec:model_merging_intro}, we select the hyperparameters from $\left\{0.01,\,0.2,\,0.4,\,0.6,\,0.8,\,1.0\right\}$.
For the scaling factor $\lambda$, Arcee's MergeKit can automatically normalize the magnitudes and self-adaptively selects the optimal scaling factor.

\section{Additional Experiment Results}
\label{appendix:addtional_results}

\subsection{Results of Individual Tasks}
\label{appendix:individual_tasks}
Here we report the performance of individual tasks for each language: Tamil in Table~\ref{tab:tam_tasks}, Telugu in Table~\ref{tab:tel_tasks}, Odia in Table~\ref{tab:ory_tasks}, Bengali in Table~\ref{tab:ben_tasks}, Tibetan in Table~\ref{tab:bod_tasks}, Uyghur~\ref{tab:uig_tasks} and Mongolian in Table~\ref{tab:mvf_tasks}.

\begin{table*}[t]
\small
\centering
\begin{tabular}{lcc}
\toprule
\textbf{Dataset} & \textbf{Dev} & \textbf{Test} \\
\midrule
MGSM & 50 & 200 \\
SIB-200 & 51 & 204 \\
Belebele & 720 & 2,880 \\
TC in MLiC-Eval & 48 & 504 \\
MRC in MLiC-Eval & 20 & 200 \\
RS in MLiC-Eval & 40 & 407 \\
Math in MLiC-Eval & 20 & 200 \\
\bottomrule
\end{tabular}
\caption{Number of instances for each language in the evaluation datasets. TC is short for text classification. MRC is short for machine reading comprehension. RS is short for response selection.}
\label{tab:data_statistics}
\end{table*}

\begin{table*}[t]
\small
\centering
\begin{tabular}{l|cc|c}
\toprule
& \textbf{SIB-200} & \textbf{Belebele} & \textbf{Average} \\
\midrule
\textbf{BASE} & 30.88 & 25.42 & 28.15 \\ 
\textbf{SFT-\texttt{flan}} & 30.88 & 27.50  & 29.19 \\
\textbf{CT-\texttt{tam}} & 73.53 & 30.83 & 52.18 \\ 
\textbf{CTSFT-\texttt{tam}-\texttt{mt}} & 69.61 & 31.53 & 50.57\\
\textbf{CTSFT-\texttt{tam}-\texttt{flan}} & 71.08 & 36.81 & 53.95 \\
\textbf{WAVG-\texttt{tam}-\texttt{flan}} & 80.39 & 34.72 & 57.56 \\
\textbf{TIES-\texttt{tam}-\texttt{flan}} & 80.39 & 36.53 & 58.46 \\
\bottomrule
\end{tabular}
\caption{The performance of different models on the Tamil tasks.}
\label{tab:tam_tasks}
\end{table*}

\begin{table*}[t]
\small
\centering
\begin{tabular}{l|ccc|c}
\toprule
& \textbf{MGSM} & \textbf{SIB-200} & \textbf{Belebele} &  \textbf{Average} \\
\midrule
\textbf{BASE} & 2.50 & 27.45 & 26.53 & 18.83 \\ 
\textbf{SFT-\texttt{flan}} & 1.00 & 24.02 & 26.81 & 17.28 \\
\textbf{CT-\texttt{tel}} & 2.00 & 73.53 & 28.47 & 34.67 \\ 
\textbf{CTSFT-\texttt{tel}-\texttt{mt}} & 3.00 & 61.27 & 34.44 & 32.90 \\
\textbf{CTSFT-\texttt{tel}-\texttt{flan}} & 7.00 & 77.45 & 29.44 & 37.96 \\
\textbf{WAVG-\texttt{tel}-\texttt{flan}} & 3.00 & 76.96 & 32.78 & 37.58 \\
\textbf{TIES-\texttt{tel}-\texttt{flan}} & 8.00 & 77.45 & 33.06 & 39.50 \\
\bottomrule
\end{tabular}
\caption{The performance of different models on the Telugu tasks.}
\label{tab:tel_tasks}
\end{table*}

\begin{table*}[t]
\small
\centering
\begin{tabular}{l|cc|c}
\toprule
& \textbf{SIB-200} & \textbf{Belebele} & \textbf{Average} \\
\midrule
\textbf{BASE} & 26.47 & 26.81 & 26.64\\ 
\textbf{SFT-\texttt{flan}} & 25.98 & 24.44 & 25.21 \\
\textbf{CT-\texttt{ory}} &  69.61 & 26.25 & 47.93 \\ 
\textbf{CTSFT-\texttt{ory}-\texttt{mt}} & 32.25 & 27.92 & 30.14 \\
\textbf{CTSFT-\texttt{ory}-\texttt{flan}} & 61.76 & 27.36 & 44.56\\
\textbf{WAVG-\texttt{ory}-\texttt{flan}} & 77.45 & 29.72 & 53.59 \\
\textbf{TIES-\texttt{ory}-\texttt{flan}} & 81.86 & 31.11 & 56.49\\
\bottomrule
\end{tabular}
\caption{The performance of different models on the Odia tasks.}
\label{tab:ory_tasks}
\end{table*}

\begin{table*}[t]
\small
\centering
\begin{tabular}{l|ccc|c}
\toprule
& \textbf{MGSM} & \textbf{SIB-200} & \textbf{Belebele} &  \textbf{Average} \\
\midrule
\textbf{BASE} & 1.00 & 50.00 & 25.42 & 25.47 \\ 
\textbf{SFT-\texttt{flan}} & 1.50 & 47.06 & 25.97 & 24.84\\
\textbf{CT-\texttt{ben}} & 1.50 & 63.73 & 27.08  & 30.77 \\ 
\textbf{CTSFT-\texttt{ben}-\texttt{mt}} & 7.00 & 76.96 & 31.25 & 38.40 \\
\textbf{CTSFT-\texttt{ben}-\texttt{flan}} &  7.50 & 80.88 & 38.19 & 42.19 \\
\textbf{WAVG-\texttt{ben}-\texttt{flan}} & 3.50 & 76.96 & 31.11 & 37.19 \\
\textbf{TIES-\texttt{ben}-\texttt{flan}} & 7.50 & 79.41 & 34.02 & 40.31 \\
\bottomrule
\end{tabular}
\caption{The performance of different models on the Bengali tasks.}
\label{tab:ben_tasks}
\end{table*}

\begin{table*}[t]
\small
\centering
\begin{tabular}{l|cccc|c}
\toprule
& \textbf{TC} & \textbf{MRC} & \textbf{RS} &  \textbf{Math} & \textbf{Average} \\
\midrule
\textbf{BASE} & \ \ 0.60 & 28.50 & 22.36 & \ \ 2.50 & 13.49\\ 
\textbf{SFT-\texttt{flan}} & 14.48 & 44.00 & 32.68 & \ \ 2.00 & 23.29\\
\textbf{CT-\texttt{bod}} & \ \ 0.40 & 28.00 & 18.18 & \ \ 7.50 & 13.52\\ 
\textbf{CTSFT-\texttt{bod}-\texttt{mt}} & 48.41 & 42.50 & 30.47 & 14.00 & 33.85\\
\textbf{CTSFT-\texttt{bod}-\texttt{flan}} & 70.24 & 46.50 & 45.70 & \ \ 7.00 & 42.36\\
\textbf{WAVG-\texttt{bod}-\texttt{flan}} & 74.40 & 51.50 & 40.79 & 10.50 & 44.30\\
\textbf{TIES-\texttt{bod}-\texttt{flan}} & 78.17 & 56.00 & 42.26 & 15.00 & 47.86 \\
\bottomrule
\end{tabular}
\caption{The performance of different models on the Tibetan tasks. All four tasks are from MLiC-Eval. TC is short for text classification. MRC is short for machine reading comprehension. RS is short for response selection.}
\label{tab:bod_tasks}
\end{table*}

\begin{table*}[t]
\small
\centering
\begin{tabular}{l|cccc|c}
\toprule
& \textbf{TC} & \textbf{MRC} & \textbf{RS} &  \textbf{Math} & \textbf{Average} \\
\midrule
\textbf{BASE} & \ \ 0.00 & 26.00 & 23.34 & \ \ 4.00 & 13.34 \\ 
\textbf{SFT-\texttt{flan}} & \ \ 7.94 & 43.00 & 34.15 & \ \ 4.00 & 22.27 \\
\textbf{CT-\texttt{uig}} & \ \ 1.39 & 21.50 & 25.31 & 11.00 & 14.80\\ 
\textbf{CTSFT-\texttt{uig}-\texttt{mt}} & 20.63 & 37.00 & 28.26 & 13.50 & 24.85 \\
\textbf{CTSFT-\texttt{uig}-\texttt{flan}} & 90.28 & 53.50 & 38.57 & 15.50 & 49.46 \\
\textbf{WAVG-\texttt{uig}-\texttt{flan}} & 54.56 & 54.00 & 42.01 & 20.00 & 42.64\\
\textbf{TIES-\texttt{uig}-\texttt{flan}} & 87.50 & 56.00 & 45.21 & 21.00 & 52.43 \\
\bottomrule
\end{tabular}
\caption{The performance of different models on the Uyghur tasks. All four tasks are from MLiC-Eval. TC is short for text classification. MRC is short for machine reading comprehension. RS is short for response selection.}
\label{tab:uig_tasks}
\end{table*}

\begin{table*}[t]
\small
\centering
\begin{tabular}{l|cccc|c}
\toprule
& \textbf{TC} & \textbf{MRC} & \textbf{RS} &  \textbf{Math} & \textbf{Average} \\
\midrule
\textbf{BASE} & \ \ 0.40 & 21.50 & 21.38 & \ \ 3.00 & 11.57 \\ 
\textbf{SFT-\texttt{flan}} & 10.52 & 34.50 & 33.42 & \ \ 1.00 & 19.86\\
\textbf{CT-\texttt{mvf}} & \ \ 0.60 & 12.5 & 27.76 & \ \ 3.50 & 11.09 \\ 
\textbf{CTSFT-\texttt{mvf}-\texttt{mt}} & 20.24 & 31.00 & 27.03 & \ \ 0.00  & 19.57 \\
\textbf{CTSFT-\texttt{mvf}-\texttt{flan}} & 40.87 & 30.00 & 25.80 & \ \ 0.50 & 24.29\\
\textbf{WAVG-\texttt{mvf}-\texttt{flan}} & 40.48 & 45.00 & 35.87 & \ \ 3.00 & 30.09\\
\textbf{TIES-\texttt{mvf}-\texttt{flan}} & 45.04 & 50.00 & 32.18 & \ \ 3.00 & 32.56\\
\bottomrule
\end{tabular}
\caption{The performance of different models on the Mongolian tasks. All four tasks are from MLiC-Eval. TC is short for text classification. MRC is short for machine reading comprehension. RS is short for response selection.}
\label{tab:mvf_tasks}
\end{table*}

\subsection{Case Study of Machine-Translated Data}
\label{appendix:case_study}
Table~\ref{tab:mt_cases} presents two examples of machine translation (MT) and human translation applied to SFT data. These examples illustrate the potential shortcomings of MT compared to human translation.

The MT output, generated by the NLLB model, often introduces irrelevant content and omits crucial information, as shown in the table's upper example. In the lower example, NLLB fails to translate several question options from English to Tibetan. These issues highlight the significant performance gap between human and machine translation.

\begin{table*}[t]
\small
\centering
\begin{tabular}{p{\columnwidth}}
\toprule
\textbf{English} \\
\textbf{\textit{Input}:} Give the rationale and then the answer. Is the following statement true?\\
``The United States Army Air Forces sent aircraft into the skies.''\\
\textbf{\textit{Output}:} The United States Army Air Forces were responsible for aerial military. The final answer: yes. \\
\midrule
\textbf{Uyghur (Translated by the NLLB Model)} \\
\textbf{\textit{Input}:}\\
\includegraphics[scale=0.17]{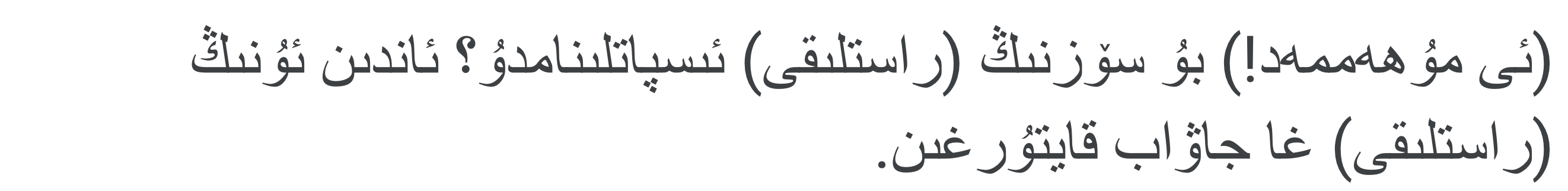}\\
\textbf{\textit{Output}:}\\
\includegraphics[scale=0.17]{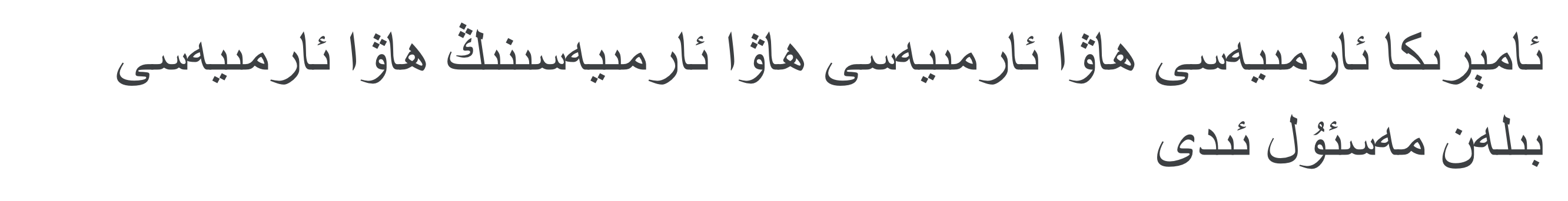}\\
\midrule
\textbf{Uyghur (Translated by Native Speakers)} \\
\textbf{\textit{Input}:}\\
\includegraphics[scale=0.17]{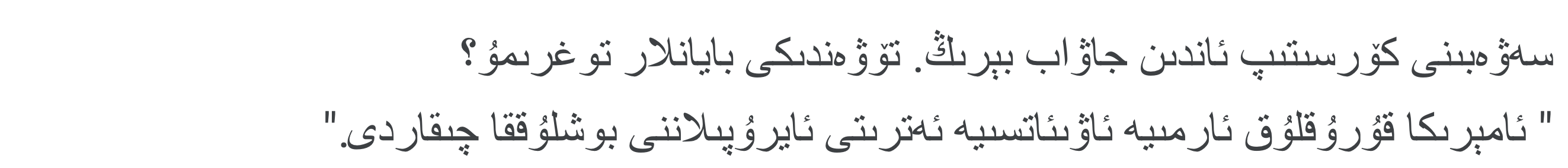}\\
\textbf{\textit{Output}:}\\
\includegraphics[scale=0.17]{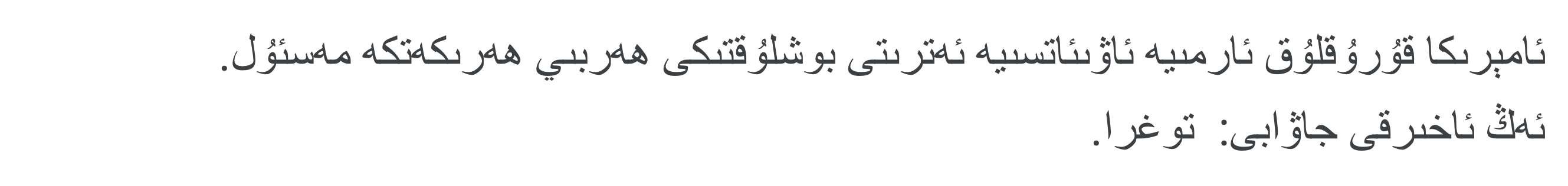}\\
\midrule
\midrule
\textbf{English} \\
\textbf{\textit{Input}:} What do people usually tie a ribbon around as a present? \\
A. floral arrangement\ \ B. girl's hair\ \ C. box\ \ \\ 
D. car \ \ E. typewriter \\ 
The best answer is\\
Answer:\\
\textbf{\textit{Output}:} A \\
\midrule
\textbf{Tibetan (Translated by the NLLB Model)} \\
\textbf{\textit{Input}:}\\
\includegraphics[scale=0.17]{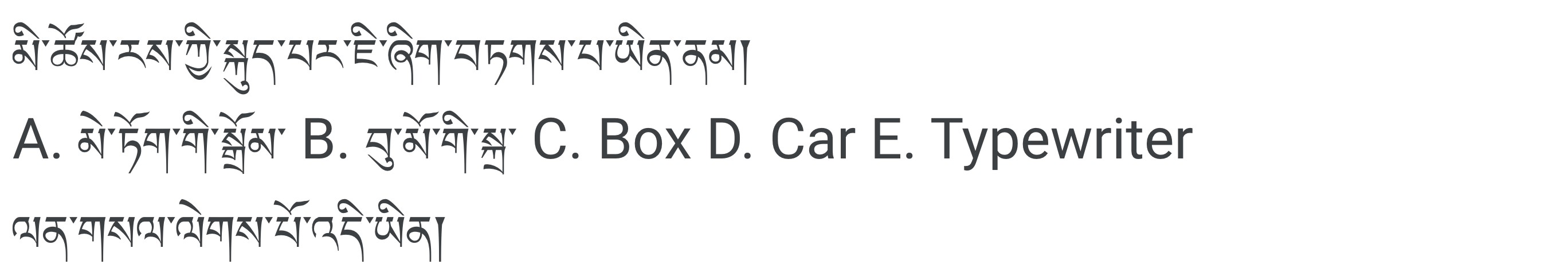}\\
\textbf{\textit{Output}:} A\\
\midrule
\textbf{Tibetan (Translated by Native Speakers)} \\
\textbf{\textit{Input}:}\\
\includegraphics[scale=0.17]{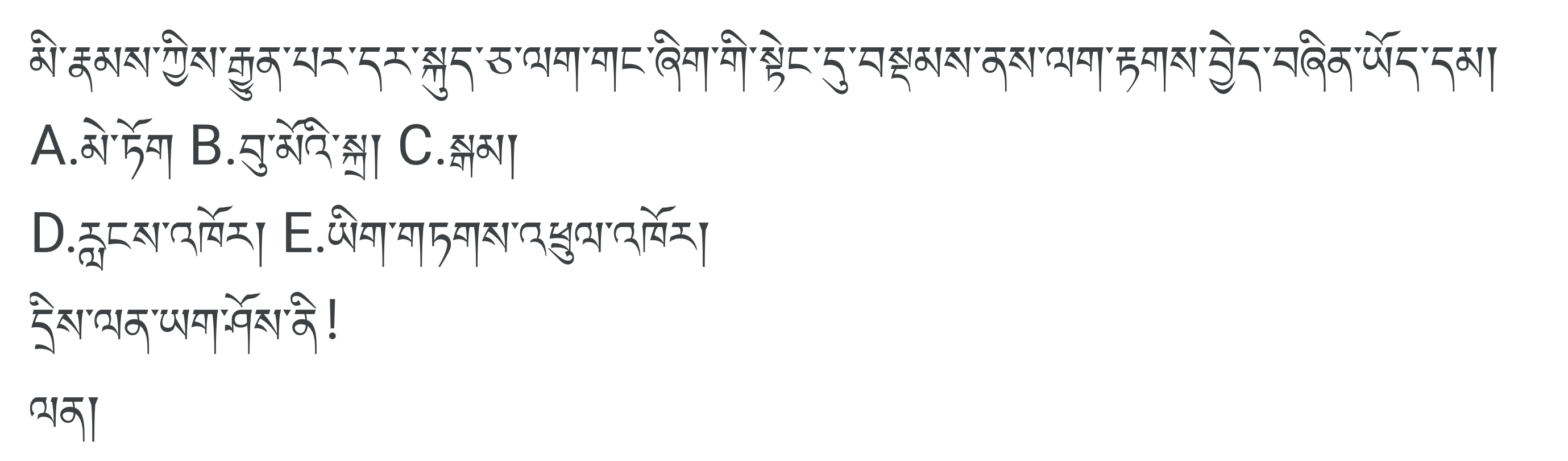}\\
\textbf{\textit{Output}:} A\\
\bottomrule
\end{tabular}
\caption{Translation samples of Uyghur (upper) and Tibetan (lower) SFT data.}
\label{tab:mt_cases}
\end{table*}

\end{document}